\pdfoutput=1
\documentclass[11pt]{article}

\usepackage[final]{acl}

\usepackage{times}
\usepackage{latexsym}
\usepackage{rotating}

\usepackage[T1]{fontenc}
\usepackage[utf8]{inputenc}
\DeclareUnicodeCharacter{2260}{$\ne$}

\usepackage{microtype}

\usepackage{inconsolata}

\usepackage{amsmath, amsfonts, amssymb}
\usepackage{algpseudocode, algorithm}
\usepackage{array}
\usepackage{graphicx}
\usepackage{multirow}
\usepackage{siunitx}
\usepackage{float}

\title{Analyzing Correlations Between Intrinsic and Extrinsic Bias Metrics of Static Word Embeddings With Their Measuring Biases Aligned}

\author{Taisei Kat\^o \and Yusuke Miyao \\
  The University of Tokyo / Tokyo, Japan \\
  \texttt{\{kato\_taisei, yusuke\}}@is.s.u-tokyo.ac.jp}

\begin{document}
\maketitle
\begin{abstract}
We examine the abilities of intrinsic bias metrics of static word embeddings to predict whether Natural Language Processing (NLP) systems exhibit biased behavior.
A word embedding is one of the fundamental NLP technologies that represents the meanings of words through real vectors, and problematically, it also learns social biases such as stereotypes.
An intrinsic bias metric measures bias by examining a characteristic of vectors, while an extrinsic bias metric checks whether an NLP system trained with a word embedding is biased.
A previous study found that a common intrinsic bias metric usually does not correlate with extrinsic bias metrics.
However, the intrinsic and extrinsic bias metrics did not measure the same bias in most cases, which makes us question whether the lack of correlation is genuine.
In this paper, we extract characteristic words from datasets of extrinsic bias metrics and analyze correlations with intrinsic bias metrics with those words to ensure both metrics measure the same bias.
We observed moderate to high correlations with some extrinsic bias metrics but little to no correlations with the others.
This result suggests that intrinsic bias metrics can predict biased behavior in particular settings but not in others.
Experiment codes are available at GitHub.\footnote{\url{https://github.com/kato8966/in-ex-correlation}}
\end{abstract}

\section{Introduction}

We re-analyze correlations between two main schemes to measure bias in static word embeddings, i.e., ``intrinsic'' and ``extrinsic'' bias metrics \citep{goldfarb-tarrant-etal-2021-intrinsic}.
A word embedding encodes stereotypical association, or bias, from a training corpus \citep{Bolukbasi_Chang_Zou_Saligrama_Kalai_2016, Caliskan_Bryson_Narayanan_2017}.
An intrinsic bias metric takes only word embeddings into account to measure bias.
On the other hand, an extrinsic bias metric trains NLP systems with word embeddings and measures the bias of word embeddings by checking if the systems behave differently to different people.
A biased NLP system works inadequately for one group of people compared to another.
To make sure that an NLP system works for everyone, we need to improve extrinsic bias metrics.

\citet{goldfarb-tarrant-etal-2021-intrinsic} found that research on reducing word embedding bias heavily relies on the assumption that lowering intrinsic bias metrics improves extrinsic bias metrics but that no research confirmed this assumption.
In response, they analyzed the relation between a common intrinsic bias metric WEAT \citep{Caliskan_Bryson_Narayanan_2017} and extrinsic bias metrics by generating and measuring the bias of word embeddings with various bias levels, from debiased ones to over-biased ones.
They found that the intrinsic and extrinsic bias metrics usually do not correlate.
They recommend researchers and practitioners not rely on intrinsic bias metrics and instead measure bias with extrinsic metrics and focus on mitigating harms caused by bias.

However, in their experiment, WEAT and extrinsic bias metrics do not measure the same bias in most cases.
For example, WEAT 7 \citep{Caliskan_Bryson_Narayanan_2017}, one of the word sets they used to measure WEAT, measures bias such as ``Boys are good at math, while girls are good at literature.''
This bias is not matched with the one of the WinoBias \citep{zhao-etal-2018-gender} extrinsic bias metric, which is like ``Doctors are men, while secretaries are women.''
Although both are about gender bias, the former is about gender bias in school subjects, while the latter is about gender bias in professions.
This mismatch makes us question whether the lack of correlation is genuine.

\citet{cao-etal-2022-intrinsic} analyzed correlations between intrinsic and extrinsic bias metrics of language models with their measuring biases matched.
They extracted characteristic words from a dataset for an extrinsic bias metric to make intrinsic and extrinsic bias metrics measure the same bias.
An intrinsic bias metric was measured with those words.
\citet{cao-etal-2022-intrinsic} measured the bias of 19 pre-trained language models with bias metrics.
However, none of the language models underwent a debiasing procedure.
It is necessary to generate many data points with various bias levels to analyze correlations between metrics.

In this study, we make intrinsic and extrinsic bias metrics measure the same bias and re-analyze their correlations, using word embeddings with various bias levels as data points.
Word embeddings are smaller and easier to train than language models, which are suitable properties to generate many data points with various bias levels.
We extract bias-representing word sets from a dataset for an extrinsic bias metric and analyze correlations with intrinsic bias metrics measured with those word sets.
For example, when analyzing correlations between intrinsic bias metrics and the WinoBias extrinsic bias metric, we extract a stereotypically male job word set, stereotypically female job word set, male word set, and female word set and measure the intrinsic bias metrics with them.

We observed moderate to high correlations with some extrinsic bias metrics but little to no correlations with the others.
This result suggests that intrinsic bias metrics can be trusted in particular settings but not in others.

\section{Bias Statement}

We use intrinsic and extrinsic bias metrics to measure bias in word embeddings.
For intrinsic bias metrics, we use WEAT and RNSB (Sections \ref{subsec:weat} and \ref{subsec:rnsb}).
We train coreference resolution models and hate speech detection models from word embeddings for extrinsic bias metrics and check if they are biased.
We call an NLP system biased if it behaves differently to different people.
For coreference resolution systems, we measure performances on stereotypical and anti-stereotypical sentences in WinoBias \cite{zhao-etal-2018-gender} (Section \ref{subsec:winobias}).
For hate speech detection systems, we compare performances between male- and female-targeting tweets \cite{goldfarb-tarrant-etal-2021-intrinsic} and white-sounding and African-American-sounding tweets (Section \ref{subsec:hsd}).

The harm extrinsic bias metrics capture is clear:
A biased coreference resolution system reinforces gender stereotypes in society and pays little attention to those with anti-stereotypical jobs.
A biased hate speech detection system unfairly censors African-American English posts \cite{davidson-etal-2019-racial, goldfarb-tarrant-etal-2021-intrinsic} and makes African-American English speakers less visible in SNS.
However, the harm intrinsic bias metrics capture is unclear, which undermines the validity of intrinsic bias metrics.
Following \citet{goldfarb-tarrant-etal-2021-intrinsic}, we conduct a correlation study between intrinsic and extrinsic bias metrics to figure out if intrinsic bias metrics are trustworthy.

\section{Background}

In this section, we describe bias metrics of which we analyze correlations and related work.

\subsection{WEAT} \label{subsec:weat}

WEAT is a common intrinsic bias metric to measure stereotypical association (e.g., ``physician'' and ``man'' or ``secretary'' and ``woman'') \citep{Caliskan_Bryson_Narayanan_2017}.
For example, suppose we want to measure gender bias in occupations.
WEAT requires word sets $T_1$ and $T_2$ ($|T_1| = |T_2|$) that represent two target concepts of bias (stereotypically male and female jobs).
They are called target words. % TK: Italic?
It also requires \textit{attribute words} $A_1$ and $A_2$ (male and female words) that are stereotypically associated with each target word set even though they should not.
``physician'' should be as close to ``woman'' as to ``man.''
Given those words, WEAT is defined as \[\sum_{t \in T_1} s(t, A_1, A_2) - \sum_{t \in T_2} s(t, A_1, A_2)\] where
\begin{align*}s(t, A_1, A_2) = &\frac{1}{|A_1|}\sum_{a \in A_1} \cos(t, a)\\&- \frac{1}{|A_2|}\sum_{a \in A_2} \cos(t, a).
\end{align*}
By normalizing the WEAT value, one can obtain the effect size, i.e., a value indicating how different target word sets are in terms of associativity with attribute words:
\[\frac{\frac{1}{|T_1|}\sum_{t \in T_1} s(t, T_1, T_2) - \frac{1}{|T_2|}\sum_{t \in T_2} s(t, A_1, A_2)}{\text{std\_dev}_{t \in T_1 \cup T_2} s(t, A_1, A_2)}.\]
\citet{Caliskan_Bryson_Narayanan_2017} also defined word sets, WEAT 1 to 10, with which WEAT is measured.

\subsection{RNSB} \label{subsec:rnsb}
The Relative Negative Sentiment Bias (RNSB) \citep{sweeney-najafian-2019-transparent} is an intrinsic bias metric to assess how much negative sentiment is associated with identity terms such as demonyms.
Let $A_1$ and $A_2$ be positive and negative sentiment word sets, respectively, and $\{t_1, ..., t_N\}$ be identity terms representing $N$ types of people.
It first trains a logistic regression sentiment classifier $f$ with $A_1$ and $A_2$.
After training, it predicts negative sentiment probability $\overline{f}(t_i)$ for each identity term $t_i$.
By normalizing $\overline{f}(t_1), ..., \overline{f}(t_N)$, we can obtain a probabilistic distribution
\begin{equation}
    D = \left(\frac{\overline{f}(t_1)}{\sum_i \overline{f}(t_i)}, ..., \frac{\overline{f}(t_N)}{{\sum_i \overline{f}(t_i)}}\right)
    \label{eq:rnsb_d}
\end{equation}
over identity terms.
No identity term should be predicted as more negative than any other term, and the distribution $D$ should be uniform.
The RNSB is the Kullback–Leibler divergence (KL divergence) of $D$ from a uniform distribution $U$ over $N$ identity terms.

\citet{wefe2020} extended the RNSB to incorporate target and attribute words.
Given attribute word sets $A_1$ and $A_2$\footnote{Note that $A_1$ and $A_2$ are no longer restricted to positive and negative sentiment words.} and target word sets $T_1, ..., T_N$, the RNSB trains logistic regression $f$ with $A_1$ as positive examples and $A_2$ as negative ones.
Then, the RNSB is the KL divergence of a probabilistic distribution over $w \in T_1 \cup ... \cup T_N$ from a uniform distribution.
However, in this extension, the distinction of the target word sets has no effect.
Even if we merge all target words and create one big target word set $T$, the RNSB is unchanged.

Instead of the probabilistic distribution over all target words, we consider a distribution over target word sets.
Let $\overline{f}(T_i)$ be the average probability that word $w$ in $T_i$ is a negative example.
We can measure the RNSB by replacing $\overline{f}(t_i)$ with $\overline{f}(T_i)$ in Equation \ref{eq:rnsb_d}.
However, we can simplify the calculation in our experiment.
We only have cases where $N = 2$, and thus how different $D$ is from the uniform distribution is represented by
\begin{align*}
    & \frac{\overline{f}(T_2)}{\overline{f}(T_1) + \overline{f}(T_2)} - \frac{\overline{f}(T_1)}{\overline{f}(T_1) + \overline{f}(T_2)}\\
    & = \frac{\overline{f}(T_2) - \overline{f}(T_1)}{\overline{f}(T_1) + \overline{f}(T_2)}.
\end{align*}
This simplifies the calculation and gives us a more fine-grained metric.
Imagine a word embedding $v$ that is overly debiased.
That is, $A_1$ is strongly associated with $T_2$, and $A_2$ is with $T_1$.
This makes $\overline{f}(T_1)$ higher than $\overline{f}(T_2)$.
The KL-divergence-based RNSB only tells us $v$ is heavily biased, while our RNSB tells us it is anti-stereotypically biased.

\subsection{WinoBias} \label{subsec:winobias}

WinoBias \citep{zhao-etal-2018-gender} is a dataset for bias evaluation of coreference resolution systems.
It contains sentences representing pro- and anti-stereotypical situations and measures the bias of the systems by comparing outputs for both situations.
\citet{goldfarb-tarrant-etal-2021-intrinsic} trained coreference resolution systems with word embeddings and let the bias scores of the systems be bias scores of the word embeddings.
In sentences in pro-stereotypical sets (e.g., ``The physician hired the secretary because he was overwhelmed with clients.''), antecedents of third-person pronouns are occupations that are stereotypically associated with the genders of the pronouns.
In anti-stereotypical sets, the genders of the pronouns are swapped.
If a coreference resolution system does not have gender bias, its outputs for both sentences should be the same.
Let $p_\text{pro}$ and $p_\text{anti}$ be performances (precision or recall) of the system on pro- and anti-stereotypical sets, respectively.
\citet{goldfarb-tarrant-etal-2021-intrinsic} defined a bias score as $p_\text{pro} - p_\text{anti}.$
WinoBias has two types, and sentences in Type 2 include syntactic cues to resolve coreferences (e.g., ``The secretary called the physician and told him about a new patient.'').

\subsection{Hate Speech Detection} \label{subsec:hsd}

\citet{goldfarb-tarrant-etal-2021-intrinsic} used hate speech detection (HSD) for extrinsic bias metric.
They extracted and annotated 15,000 tweets as targeting male, female, or neutral from the abusive tweet dataset of \citet{founta-etal-2022}.
Letting $p_\text{male}$ and $p_\text{female}$ be performances of a hate speech detector on tweets targeting male and female, respectively, they defined a bias score of a word embedding as $p_\text{male} - p_\text{female}.$
In addition to that, we measure the difference $p_\text{w} - p_\text{aa}$ between performances for white-sounding tweets $p_\text{w}$ and African-American-sounding tweets $p_\text{aa}$.
African-American English is often misunderstood as hateful \citep{sap-etal-2019-risk}, and therefore, comparing the HSD performances is a good extrinsic bias metric for racial bias.

\subsection{Related Work}

\citet{goldfarb-tarrant-etal-2021-intrinsic} found that research on reducing word embedding bias heavily relies on the assumption that reducing intrinsic bias improves extrinsic bias metrics but that no research confirmed this.
To test this assumption, they analyzed the relation between the WEAT effect size and extrinsic bias metrics under various conditions (two word embedding algorithms, two downstream tasks, and two languages, English and Spanish).
They generated word embeddings with various bias levels, measured the biases with WEAT and extrinsic bias metrics, and analyzed their relations by calculating Pearson's $r$ and making scatter plots.
As a result of the experiment, \citet{goldfarb-tarrant-etal-2021-intrinsic} found that WEAT does not correlate with extrinsic bias metrics in most cases.
However, the biases that WEAT measures and the ones that extrinsic bias metrics measure were not aligned well, which makes us question whether the lack of correlation is genuine.
In our study, we close this gap and re-analyze correlations.

For language models, \citet{cao-etal-2022-intrinsic} analyzed correlations between intrinsic and extrinsic bias metrics with adjustments of several misalignments.
For example, when measuring correlations with toxicity-related extrinsic bias metrics, they manually chose toxic and anti-toxic words from word clouds generated from extrinsic bias metric dataset Jigsaw Toxicity \citep{jigsaw-unintended-bias-in-toxicity-classification}, and measured CEAT \citep{Guo_Caliskan_2021} with them.
However, their analysis is based on 19 pre-trained language models, none of which underwent a debiasing procedure.
It is necessary to generate many data points with various degrees of bias to analyze correlations between metrics.
In our study, we generate many word embeddings with various bias levels, following \citet{goldfarb-tarrant-etal-2021-intrinsic}, and analyze correlations.
Even though it is less potent than a large language model, a word embedding is smaller and easier to train.
It is suitable for generating many data points with various bias levels.

\citet{orgad-etal-2022-gender} also examined the relationship between intrinsic and extrinsic bias metrics of language models.
They proposed a new intrinsic bias metric, ``compression,'' and showed it correlates well with extrinsic bias metrics to some degree.
However, they considered only gender bias.
We take racial bias into account as well.

\section{Method}

Given a dataset to measure an extrinsic bias metric, we extract target and attribute word sets from it that represent bias the extrinsic bias metric measures.
By feeding those word sets to an intrinsic bias metric, we can make it measure the same bias as the extrinsic bias metric.
Then, we analyze the correlation between the metrics, following \citet{goldfarb-tarrant-etal-2021-intrinsic}.
Ensuring that both metrics measure the same bias is essential to analyze the correlation accurately.
Suppose it does not hold, and the correlation is weak; we cannot conclude whether it is due to a misalignment of biases measured by both metrics or a fundamental difference between them.

From WinoBias, we extract stereotypical male/female occupations as target words and $\{\text{``he''}, \text{``him''}\}$ and $\{\text{``she''}, \text{``her''}\}$ as attribute words.\footnote{Stereotypical male occupations contain a two-word expression (i.e., ``construction worker''). Its word embedding is an average of ``construction'' and ``worker.''}

An HSD extrinsic bias metric measures bias by comparing performances for two groups (in our experiment, male-targeting tweets vs. female-targeting tweets \citep{goldfarb-tarrant-etal-2021-intrinsic} and white-sounding tweets vs. African-American-sounding tweets).
Its measuring bias is that one group is associated with hatefulness.
We extract words representing the two groups as target words and hateful and unhateful words as attribute words.
We extract those words based on Pointwise Mutual Information (PMI) \citep{Fano_1963}.
Given an HSD dataset consisting of tweets with two labels (one indicating hatefulness and the other showing a group of people), we first calculate PMIs between an event where a tweet is labeled as $l \in \{\text{HS}, \text{NON-HS}\}$ (HS stands for ``hate speech.'') and an event where the word $w$ appears in a tweet.
Let it be $\text{PMI}(l, w)$.
To calculate $\text{PMI}(l, w)$, we assume the Na\"ive Bayes generative model, where a tweet is labeled as $l$ with a probability of $\mu_l$ and its tokens are sampled from a categorical distribution parameterized with $\boldsymbol{\theta}_l$.
After estimating the parameters $\boldsymbol\mu$ and $\boldsymbol\theta$, we calculate $\text{PMI}(l, w)$ as
\[\log_2 \frac{\theta_{l, w}}{\sum_{l^\prime}\mu_{l^\prime} \theta_{l^\prime, w}}.\]
We extract candidate hateful attribute words $w_1, ..., w_{40}$ where $\text{PMI}(\text{HS}, w_i)$ is the $i$-th largest.
We do not want to pick up rare words.
Therefore, we filter out words appearing in less than ten tweets in the dataset, and when candidate attribute words have the same PMI, the more frequent word is chosen.
We then manually select hateful attribute words from the candidates.
Non-hateful attribute words and target words are extracted in similar ways.
(When extracting, for example, men and women target words, let $l \in \{\text{MALE}, \text{FEMALE}\}$.)
A detailed selection process is explained in Appendix~\ref{app:word_ext}.

\section{Experiment}

Following \citet{goldfarb-tarrant-etal-2021-intrinsic}, we generate word embeddings with various bias levels, measure their bias with intrinsic and extrinsic bias metrics, and analyze their correlations.
We use the FUJITSU Server PRIMERGY GX2570 M6 in the Information Technology Center at the University of Tokyo to conduct experiments.

\subsection{Bias Metrics} \label{subsec:bias_metrics}

We employ the RNSB (intrinsic bias metric) and the racial performance gap of HSD (extrinsic bias metric), in addition to metrics used in the English experiment of \citet{goldfarb-tarrant-etal-2021-intrinsic} (the WEAT effect size, WinoBias, and the gender performance gap of HSD).
The RNSB is similar to extrinsic bias metrics in that both train a machine learning model.
We include the RNSB because it might correlate well with extrinsic bias metrics.
The RNSB is also similar to the intrinsic bias metric ``compression'' \citep{orgad-etal-2022-gender} in that both measure bias with classifiers.
Details about measuring the RNSB are in Appendix~\ref{app:rnsb}.

To measure the racial performance gap, we need to generate white-sounding and African-American-sounding tweet sets.
We use the mixed-membership demographic-language model developed by \citet{blodgett-etal-2016-demographic}.
It receives a list of word tokens and predicts the proportions of dialects of each race and ethnicity (white, African-American, Hispanic, and Asian).
We take the argmax of these proportions for each tweet in the test dataset to determine race or ethnicity.

For extrinsic bias metrics, \citet{goldfarb-tarrant-etal-2021-intrinsic} compared precisions and recalls.
We also compare the F1 scores to see the overall performance difference.

NLP models trained for the extrinsic bias metrics are the same as in \citet{goldfarb-tarrant-etal-2021-intrinsic}.
The coreference resolution model is proposed by \citet{lee-etal-2017-end}, a high-performance, end-to-end neural model.
We use \citet{goldfarb-tarrant-etal-2021-intrinsic}'s implementation.
It is trained with OntoNotes \citep{Weischedel2017OntoNotesA} 5.0, split into training/validation/test datasets as in CoNLL-2012 Shared Task \citep{pradhan-etal-2012-conll}.
The HSD model is based on the ``CNN-static'' of \citet{kim-2014-convolutional}, a simple text classifier applying several convolutional layers in parallel.
Hyperparameter values are in Appendix~\ref{app:hsd}.
It is trained with the abusive tweet dataset of \citet{founta-etal-2022}.
Dataset preprocessing is explained in Appendix~\ref{app:hsd_dataset}.
Basic statistics of the datasets are shown in Appendix~\ref{app:stats}.

We measure intrinsic bias metrics with the extracted words and WEAT words to see the influence of word sets.
When measuring gender bias, we use WEAT 6, 7, and 8, which were used in \citet{goldfarb-tarrant-etal-2021-intrinsic}.
We use WEAT 3, 4, and 5 to measure racial bias.
\citet{goldfarb-tarrant-etal-2021-intrinsic} substituted general male and female words for male and female names in WEAT 6 because these names do not appear in WinoBias.
We adopt the substitution.
Since our training corpora of word embeddings are all lowercased, we lowercase WEAT words when measuring intrinsic bias metrics.
Some target words in WEAT word sets do not appear in a training corpus of word embeddings.
We remove those words and an equal number of words from the other target word set so that both target word sets have the same number of words.
Appendix~\ref{app:weat} shows WEAT words used in our experiment.

\begin{table*}[ht]
    \centering
    \begin{tabular}{lp{3.2cm}p{3.2cm}cc}
        \hline
        Name & $T_1$ & $T_2$ & $A_1$ & $A_2$\\
        \hline
        WinoBias word sets & Stereotypical male jobs & Stereotypical female jobs & Male & Female\\
        HSD gender bias word sets & Male & Female & Hateful & Unhateful\\
        HSD racial bias word sets & White-sounding words & African-American-sounding words & Unhateful & Hateful\\
        WEAT 3, 4, and 5 & European American names & African American names & Pleasant & Unpleasant\\
        WEAT 6 & Career & Family & Male & Female\\
        WEAT 7 & Math & Arts & Male & Female\\
        WEAT 8 & Science & Arts & Male & Female\\
        \hline
    \end{tabular}
    \caption{Summarization of word sets we used. $T_1$ and $T_2$ are target words, and $A_1$ and $A_2$ are attribute words. The original WEAT 6 has male and female target words and career and family attribute words. We swap them, as \citet{goldfarb-tarrant-etal-2021-intrinsic} did so that male and female words are attribute words as in WEAT 7 and 8.} \label{tab:wordset}
\end{table*}
Table~\ref{tab:wordset} summarizes the word sets used in our experiment.

\subsection{Training of Word Embeddings}

Training settings for word embeddings are similar to \citet{goldfarb-tarrant-etal-2021-intrinsic}.
We train Skip-gram word2vec with negative sampling \citep{Mikolov_Sutskever_Chen_Corrado_Dean_2013} and fastText \citep{bojanowski-etal-2017-enriching} word embeddings.
Word embeddings for coreference resolution are trained on Wikipedia and on Twitter data for HSD.
A detailed explanation is given in Appendix~\ref{app:word_emb}.

\subsection{Bias Modification}

To modify bias, we need word sets to define bias as in intrinsic bias metrics.
\citet{goldfarb-tarrant-etal-2021-intrinsic} obtained these word sets by merging word sets with which intrinsic bias metrics were measured, expanding them following \citet{Lauscher_Glavaš_Ponzetto_Vulić_2020}, and removing inappropriate words manually.
Given a word embedding $v$ (not an embedding of which we measure bias), merged target word sets $T_1$ and $T_2$, and merged attribute word sets $A_1$ and $A_2$, for each word $w$ in a word set, its $K$ closest words ${\{w_i\}}_i$ in $v$ in cosine similarity terms are added to the word set.
If $w_i$ is already in $T_1 \cup T_2 \cup A_1 \cup A_2$, it is skipped.
Details about the hyperparameters we chose and removing inappropriate words are in Appendix~\ref{app:bias_modi_wordsets}.
Table~\ref{tab:bias_modification_eval_wordsets} summarizes with which word sets word embeddings are modified and evaluated.

\begin{table*}
\centering
\begin{tabular}{cc}
\hline
Modification word sets & Evaluation word sets \\
\hline
\multirow{3}{*}{Expansion of merged WEAT 6, 7, and 8} & WEAT 6 \\
 & WEAT 7 \\
 & WEAT 8 \\
\hline
Expansion of WinoBias word sets & WinoBias word sets \\
\hline
Expansion of HSD gender bias word sets & HSD gender bias word sets \\
\hline
\multirow{3}{*}{Expansion of merged WEAT 3, 4, and 5} & WEAT 3 \\
 & WEAT 4 \\
 & WEAT 5 \\
\hline
Expansion of HSD racial bias word sets & HSD racial bias word sets \\
\hline
\end{tabular}
    \caption{Summarization of word sets with which word embeddings are modified and evaluated.} \label{tab:bias_modification_eval_wordsets}
\end{table*}

Following \citet{goldfarb-tarrant-etal-2021-intrinsic}, we ran the dataset balancing \citep{Dixon_Li_Sorensen_Thain_Vasserman_2018} and \textsc{Attract-Repel} \citep{mrksic-etal-2017-semantic} to generate word embeddings with various bias levels.
They ran the algorithms several times until certain conditions were met.
We create word embeddings with various degrees of bias by changing the parameters of the algorithms.
This method allows us to create more word embeddings.
In total, 45 word embeddings are generated.
This is more than twice as many data points as in \citet{goldfarb-tarrant-etal-2021-intrinsic} and \citet{cao-etal-2022-intrinsic} (which analyzed correlations for language models).

The dataset balancing makes a dataset more balanced (e.g., in terms of frequencies of male and female entities) so that a model trained on it is less biased.
\citet{goldfarb-tarrant-etal-2021-intrinsic} reduced stereotypical sentences (e.g., ``He is a physician.'') to reduce bias and anti-stereotypical sentences to enhance bias.
Here, a stereotypical sentence contains words in $T_1$ and $A_1$ or $T_2$ and $A_2$.
($T_1$ and $T_2$ are target word sets, and $A_1$ and $A_2$ are attribute word sets.)
An anti-stereotypical sentence contains words in $T_1$ and $A_2$ or $T_2$ and $A_1$.
When running dataset balancing on the Twitter corpus, we regard a tweet as a sentence.
We introduce a parameter $p \in [0, 1)$ (a sampling probability) to the algorithm.
For each (anti-)stereotypical sentence, we keep it with probability $p$ and discard it with probability $1 - p$.
We run the algorithm with $p \in \{0.0, 0.1, 0.2, ..., 0.9\}$ for both debiasing and over-biasing.

\textsc{Attract-Repel} is an algorithm designed to improve the semantics of word embeddings.
It makes synonym pairs closer and antonym pairs far away.
\citet{goldfarb-tarrant-etal-2021-intrinsic} re-purposed it for bias modification.
Given target words $T_1$ and $T_2$ and attribute words $A_1$ and $A_2$, they debiased word embeddings by setting $T_1 \times A_2 \cup T_2 \times A_1$ as synonym pairs and $T_1 \times A_1 \cup T_2 \times A_2$ as antonym pairs.
When over-biasing, synonym and antonym pairs were swapped.
We run the algorithm while changing the parameters.
The combinations of the parameters are in Appendix~\ref{app:ar_paras}.

\section{Results and Discussion} \label{sec:results_dis}

\begin{figure*}[ht]
    \centering
    \includegraphics[width=\linewidth]{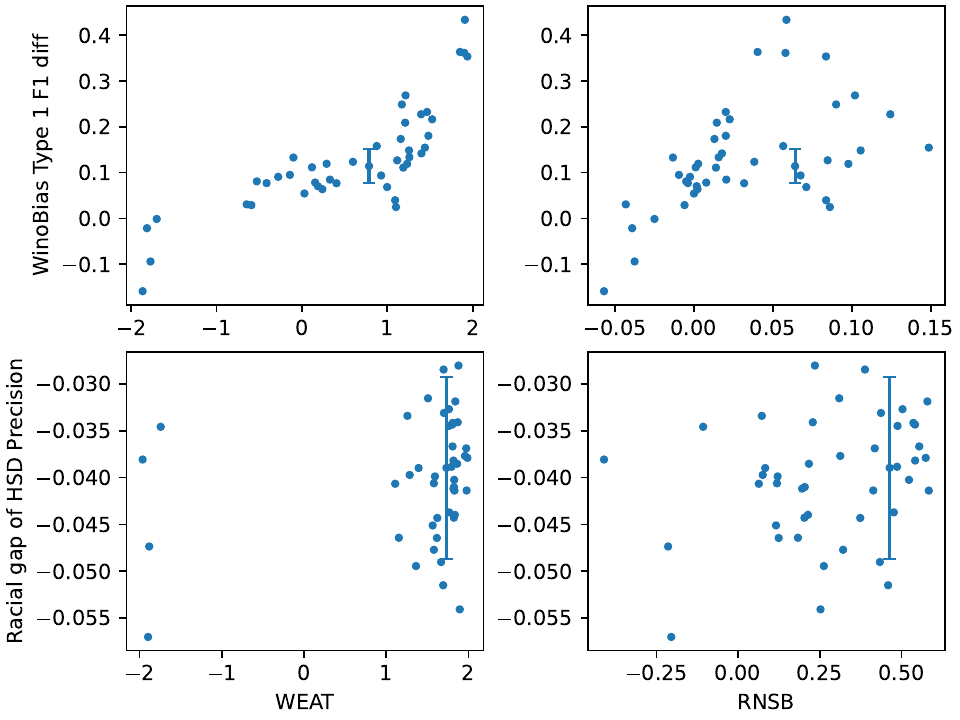}
    \caption{Correlations between intrinsic bias metrics (x-axis) and extrinsic bias metrics (y-axis). The intrinsic bias metrics are measured with the corresponding extracted word sets. The word embedding algorithm used is word2vec. For original (not bias-modified) word embeddings, we train NLP models ten times so that we can calculate standard deviations of extrinsic bias metrics, which are shown as error bars.}
    \label{fig:results}
\end{figure*}

Table~\ref{tab:results_gender_w2v}, \ref{tab:results_gender_ft}, \ref{tab:results_race_w2v}, and \ref{tab:results_race_ft} show Spearman's correlations between intrinsic and extrinsic bias metrics.
Table~\ref{tab:results_gender_w2v_pv}, \ref{tab:results_gender_ft_pv}, \ref{tab:results_race_w2v_pv}, and \ref{tab:results_race_ft_pv} show p-values calculated by two-sided permutation test.
Figure~\ref{fig:results} shows a couple of scatter plots.\footnote{Refer to the supplementary material for all the scatter plots.}

\begin{table*}
\centering
\begin{tabular}{|l l | S  S | S  S | S  S | S  S|}
\hline
&  & \multicolumn{8}{l|}{Word sets with which intrinsic bias metrics are measured} \\
\hline
& & \multicolumn{2}{l|}{Extracted} & \multicolumn{2}{l|}{WEAT 6} & \multicolumn{2}{l|}{WEAT 7} & \multicolumn{2}{l|}{WEAT 8} \\
\multicolumn{2}{|l|}{Extrinsic bias metric} & {WEAT} & {RNSB} & {WEAT} & {RNSB} & {WEAT} & {RNSB} & {WEAT} & {RNSB} \\
\hline
\multirow{3}{0.93cm}{WB type 1} & Precision diff & 0.84 & 0.56 & 0.09 & 0.08 & 0.11 & 0.11 & 0.07 & 0.07 \\
 & Recall diff & 0.84 & 0.58 & 0.04 & 0.05 & 0.07 & 0.08 & 0.03 & 0.04 \\
 & F1 diff & 0.83 & 0.56 & 0.07 & 0.06 & 0.1 & 0.1 & 0.06 & 0.06 \\
\hline
\multirow{3}{0.93cm}{WB type 2} & Precision diff & 0.91 & 0.6 & -0.11 & -0.1 & -0.13 & -0.1 & -0.16 & -0.11 \\
 & Recall diff & 0.9 & 0.59 & -0.09 & -0.07 & -0.1 & -0.05 & -0.14 & -0.05 \\
 & F1 diff & 0.91 & 0.6 & -0.09 & -0.07 & -0.11 & -0.06 & -0.14 & -0.07 \\
\hline
\multirow{3}{0.93cm}{HSD} & Precision diff & -0.11 & -0.17 & -0.03 & 0.04 & -0.08 & 0.04 & -0.07 & 0.01 \\
 & Recall diff & 0.33 & 0.26 & -0.15 & -0.13 & -0.08 & -0.13 & -0.14 & -0.17 \\
 & F1 diff & 0.21 & 0.14 & -0.11 & -0.08 & -0.08 & -0.08 & -0.12 & -0.13 \\
\hline
\end{tabular}
\caption{Gender bias results with word2vec. Each cell represents Spearman's correlation between corresponding intrinsic and extrinsic bias metrics. WB stands for WinoBias.} \label{tab:results_gender_w2v}
\end{table*}

\begin{table*}
\centering
\begin{tabular}{|l l | S  S | S  S | S  S | S  S|}
\hline
&  & \multicolumn{8}{l|}{Word sets with which intrinsic bias metrics are measured} \\
\hline
& & \multicolumn{2}{l|}{Extracted} & \multicolumn{2}{l|}{WEAT 6} & \multicolumn{2}{l|}{WEAT 7} & \multicolumn{2}{l|}{WEAT 8} \\
\multicolumn{2}{|l|}{Extrinsic bias metric} & {WEAT} & {RNSB} & {WEAT} & {RNSB} & {WEAT} & {RNSB} & {WEAT} & {RNSB} \\
\hline
\multirow{3}{0.93cm}{WB type 1} & Precision diff & 0.91& 0.66& -0.1& -0.04& -0.08& -0.08& -0.1& -0.1\\
 & Recall diff & 0.91& 0.67& -0.09& 0& -0.1& -0.04& -0.11& -0.06\\
 & F1 diff & 0.91& 0.66& -0.1& -0.03& -0.1& -0.06& -0.12& -0.08\\
\hline
\multirow{3}{0.93cm}{WB type 2} & Precision diff & 0.88& 0.57& -0.17& -0.13& -0.21& -0.21& -0.2& -0.16\\
 & Recall diff & 0.89& 0.58& -0.2& -0.09& -0.22& -0.17& -0.21& -0.14\\
 & F1 diff & 0.89& 0.58& -0.17& -0.09& -0.2& -0.17& -0.19& -0.13\\
\hline
\multirow{3}{0.93cm}{HSD} & Precision diff & -0.14& -0.08& -0.23& -0.09& -0.19& -0.07& -0.25& -0.1\\
 & Recall diff & 0.21& 0.19& 0.22& 0.06& 0.24& 0.04& 0.23& 0.01\\
 & F1 diff & 0.01& 0.05& 0& -0.07& 0.03& -0.05& 0.01& -0.07\\
\hline
\end{tabular}
\caption{Gender bias results with fastText} \label{tab:results_gender_ft}
\end{table*}

\begin{table*}
\centering
\begin{tabular}{|l l | S  S | S  S | S  S | S  S|}
\hline
&  & \multicolumn{8}{l|}{Word sets with which intrinsic bias metrics are measured} \\
\hline
& & \multicolumn{2}{l|}{Extracted} & \multicolumn{2}{l|}{WEAT 3} & \multicolumn{2}{l|}{WEAT 4} & \multicolumn{2}{l|}{WEAT 5} \\
\multicolumn{2}{|l|}{Extrinsic bias metric} & {WEAT} & {RNSB} & {WEAT} & {RNSB} & {WEAT} & {RNSB} & {WEAT} & {RNSB} \\
\hline
\multirow{3}{0.93cm}{HSD} & Precision diff & 0.24& 0.29& -0.03& 0.13& -0.06& 0.16& -0.07& 0.24\\
 & Recall diff & -0.2& -0.15& 0.02& -0.06& 0.02& -0.07& 0.06& -0.01\\
 & F1 diff & -0.04& 0.12& -0.04& -0.05& -0.05& -0.05& -0.03& 0.04\\
\hline
\end{tabular}
\caption{Racial bias results with word2vec} \label{tab:results_race_w2v}
\end{table*}

\begin{table*}
\centering
\begin{tabular}{|l l | S  S | S  S | S  S | S  S|}
\hline
&  & \multicolumn{8}{l|}{Word sets with which intrinsic bias metrics are measured} \\
\hline
& & \multicolumn{2}{|l|}{Extracted} & \multicolumn{2}{l|}{WEAT 3} & \multicolumn{2}{l|}{WEAT 4} & \multicolumn{2}{l|}{WEAT 5} \\
\multicolumn{2}{|l|}{Extrinsic bias metric} & {WEAT} & {RNSB} & {WEAT} & {RNSB} & {WEAT} & {RNSB} & {WEAT} & {RNSB} \\
\hline
\multirow{3}{0.93cm}{HSD} & Precision diff & 0.09 & 0.12 & -0.17 & -0.01 & -0.17 & 0 & -0.12 & 0.14\\
 & Recall diff & -0.13 & 0.05 & 0.08 & -0.02 & 0.08 & 0 & 0.03 & -0.01\\
 & F1 diff & -0.23 & 0.12 & -0.01 & -0.03 & -0.02 & 0.03 & -0.03 & 0.12\\
\hline
\end{tabular}
\caption{Racial bias results with fastText} \label{tab:results_race_ft}
\end{table*}

\begin{table*}
\centering
\begin{tabular}{|l l | S  S | S  S | S  S | S  S|}
\hline
&  & \multicolumn{8}{l|}{Word sets with which intrinsic bias metrics are measured} \\
\hline
& & \multicolumn{2}{l|}{Extracted} & \multicolumn{2}{l|}{WEAT 6} & \multicolumn{2}{l|}{WEAT 7} & \multicolumn{2}{l|}{WEAT 8} \\
\multicolumn{2}{|l|}{Extrinsic bias metric} & {WEAT} & {RNSB} & {WEAT} & {RNSB} & {WEAT} & {RNSB} & {WEAT} & {RNSB} \\
\hline
\multirow{3}{0.93cm}{WB type 1} & Precision diff & {2e-4}& {2e-4}& 0.55& 0.63& 0.46& 0.47& 0.63& 0.65\\
 & Recall diff & {2e-4}& {2e-4}& 0.78& 0.76& 0.65& 0.59& 0.83& 0.78\\
 & F1 diff & {2e-4}& {2e-4}& 0.64& 0.68& 0.53& 0.51& 0.7& 0.7\\
\hline
\multirow{3}{0.93cm}{WB type 2} & Precision diff & {2e-4}& {2e-4}& 0.47& 0.51& 0.41& 0.52& 0.3& 0.48\\
 & Recall diff & {2e-4}& {2e-4}& 0.58& 0.67& 0.5& 0.76& 0.38& 0.73\\
 & F1 diff & {2e-4}& {2e-4}& 0.55& 0.66& 0.49& 0.7& 0.38& 0.66\\
\hline
\multirow{3}{0.93cm}{HSD} & Precision diff & 0.46& 0.26& 0.83& 0.79& 0.6& 0.8& 0.62& 0.97\\
 & Recall diff & 0.021& 0.077& 0.33& 0.41& 0.61& 0.42& 0.37& 0.27\\
 & F1 diff & 0.16& 0.38& 0.45& 0.59& 0.58& 0.59& 0.42& 0.4\\
\hline
\end{tabular}
\caption{Gender bias results with word2vec (p-values)} \label{tab:results_gender_w2v_pv}
\end{table*}

\begin{table*}
\centering
\begin{tabular}{|l l | S  S | S  S | S  S | S  S|}
\hline
&  & \multicolumn{8}{l|}{Word sets with which intrinsic bias metrics are measured} \\
\hline
& & \multicolumn{2}{l|}{Extracted} & \multicolumn{2}{l|}{WEAT 6} & \multicolumn{2}{l|}{WEAT 7} & \multicolumn{2}{l|}{WEAT 8} \\
\multicolumn{2}{|l|}{Extrinsic bias metric} & {WEAT} & {RNSB} & {WEAT} & {RNSB} & {WEAT} & {RNSB} & {WEAT} & {RNSB} \\
\hline
\multirow{3}{0.93cm}{WB type 1} & Precision diff & {2e-4}& {2e-4}& 0.53& 0.75& 0.62& 0.59& 0.52& 0.52\\
 & Recall diff & {2e-4}& {2e-4}& 0.55& 0.95& 0.51& 0.79& 0.46& 0.69\\
 & F1 diff & {2e-4}& {2e-4}& 0.48& 0.84& 0.5& 0.69& 0.42& 0.59\\
\hline
\multirow{3}{0.93cm}{WB type 2} & Precision diff & {2e-4}& {2e-4}& 0.25& 0.41& 0.16& 0.16& 0.19& 0.28\\
 & Recall diff & {2e-4}& {2e-4}& 0.2& 0.54& 0.13& 0.26& 0.16& 0.36\\
 & F1 diff & {2e-4}& {2e-4}& 0.26& 0.56& 0.19& 0.26& 0.21& 0.39\\
\hline
\multirow{3}{0.93cm}{HSD} & Precision diff & 0.36& 0.6& 0.13& 0.55& 0.21& 0.66& 0.099& 0.53\\
 & Recall diff & 0.17& 0.21& 0.15& 0.69& 0.12& 0.79& 0.14& 0.96\\
 & F1 diff & 0.93& 0.74& 0.97& 0.67& 0.87& 0.73& 0.97& 0.64\\
\hline
\end{tabular}
\caption{Gender bias results with fastText (p-values)} \label{tab:results_gender_ft_pv}
\end{table*}

\begin{table*}
\centering
\begin{tabular}{|l l | S  S | S  S | S  S | S  S|}
\hline
&  & \multicolumn{8}{l|}{Word sets with which intrinsic bias metrics are measured} \\
\hline
& & \multicolumn{2}{l|}{Extracted} & \multicolumn{2}{l|}{WEAT 3} & \multicolumn{2}{l|}{WEAT 4} & \multicolumn{2}{l|}{WEAT 5} \\
\multicolumn{2}{|l|}{Extrinsic bias metric} & {WEAT} & {RNSB} & {WEAT} & {RNSB} & {WEAT} & {RNSB} & {WEAT} & {RNSB} \\
\hline
\multirow{3}{0.93cm}{HSD} & Precision diff & 0.11& 0.051& 0.84& 0.39& 0.68& 0.27& 0.66& 0.11\\
 & Recall diff & 0.19& 0.32& 0.9& 0.7& 0.89& 0.66& 0.71& 0.95\\
 & F1 diff & 0.82& 0.43& 0.82& 0.76& 0.73& 0.74& 0.85& 0.82\\
\hline
\end{tabular}
\caption{Racial bias results with word2vec (p-values)} \label{tab:results_race_w2v_pv}
\end{table*}

\begin{table*}
\centering
\begin{tabular}{|l l | S  S | S  S | S  S | S  S|}
\hline
&  & \multicolumn{8}{l|}{Word sets with which intrinsic bias metrics are measured} \\
\hline
& & \multicolumn{2}{|l|}{Extracted} & \multicolumn{2}{l|}{WEAT 3} & \multicolumn{2}{l|}{WEAT 4} & \multicolumn{2}{l|}{WEAT 5} \\
\multicolumn{2}{|l|}{Extrinsic bias metric} & {WEAT} & {RNSB} & {WEAT} & {RNSB} & {WEAT} & {RNSB} & {WEAT} & {RNSB} \\
\hline
\multirow{3}{0.93cm}{HSD} & Precision diff & 0.56& 0.43& 0.29& 0.99& 0.26& 0.97& 0.45& 0.34\\
 & Recall diff & 0.4& 0.73& 0.59& 0.89& 0.62& 0.98& 0.83& 0.94\\
 & F1 diff & 0.14& 0.43& 0.97& 0.85& 0.93& 0.86& 0.84& 0.44\\
\hline
\end{tabular}
\caption{Racial bias results with fastText (p-values)} \label{tab:results_race_ft_pv}
\end{table*}

When WEAT is measured with the WinoBias word sets, it shows strong correlations with WinoBias contrary to the previous study \citep{goldfarb-tarrant-etal-2021-intrinsic}.
It means WEAT can roughly predict the biased behavior of coreference resolution systems.
However, when we use the WEAT word sets, which were used in \citet{goldfarb-tarrant-etal-2021-intrinsic}, the correlation disappears.
This result shows the importance of matching measuring biases of intrinsic and extrinsic bias metrics.
A similar result for language models was reported in \citet{cao-etal-2022-intrinsic}.

The RNSB shows moderate correlations with WinoBias.
However, these correlations are unreliable for estimating the WinoBias metric from the RNSB because the WinoBias metric varies greatly even for a similar RNSB value (the top-right of Figure~\ref{fig:results}).

Correlations with the HSD extrinsic bias metrics are small.
Unsurprisingly, the F1 difference has small correlations with intrinsic bias metrics.
For example, when racial intrinsic bias metrics increase, that means associations between white-sounding words and unhateful ones and between African-American-sounding words and hateful ones increase.
This makes the HSD system predict white-sounding tweets as unhateful and African-American-sounding ones as hateful.
That results in increase of precision difference $p_\text{w} - p_\text{aa}$, decrease of recall difference $r_\text{w} - r_\text{aa}$, and relatively stable F1 difference $f_\text{w} - f_\text{aa}$.

Following this logic, positive correlations of precision difference and negative correlations of recall difference are expected for racial bias.
For gender bias, the opposite trend is expected since hateful words are associated with male terms, and unhateful words are with female terms.
(Compare the HSD gender bias word sets and racial bias word sets in Table~\ref{tab:wordset}.)

The results do not match our expectations well.
The correlations are small.
The influence of word sets is also unexpected.
We observed stronger correlations for gender and racial bias in the word2vec setting when intrinsic bias metrics were measured with the extracted word sets.
However, WEAT for gender bias in the fastText setting shows the opposite trend.

One possible hypothesis for the lack of correlation is that there is a fundamental difference between intrinsic bias metrics and the HSD bias metrics.
In the HSD, an occurrence of a hateful word does not necessarily mean a tweet is hate speech.
Intrinsic bias metrics, on the other hand, focus on only word embeddings of given words (e.g., hateful words).
This difference might prevent us from matching their measuring biases.
This hypothesis also explains why the WinoBias metric strongly correlates with intrinsic bias metrics.
The WinoBias metric focuses on target and attribute words, similar to intrinsic bias metrics, and observes which target word is linked to an attribute word.

\section{Conclusions}

In this study, we re-analyzed \citet{goldfarb-tarrant-etal-2021-intrinsic} by making intrinsic and extrinsic bias metrics measure the same bias.
We extracted target and attribute words from extrinsic bias metrics and measured intrinsic bias metrics with them.
We observed moderate to high correlations with the WinoBias extrinsic bias metrics in the experiment but little to no correlations with the HSD metrics.
This result suggests that intrinsic bias metrics can predict a biased behavior in coreference resolution but not in HSD.

We focused on static word embeddings to reproduce \citet{goldfarb-tarrant-etal-2021-intrinsic}, and our work cannot directly be applied to large language models, which are mainstream foundation models in today's NLP.
Nonetheless, our work is a cautionary tale that sometimes, a small difference in an experiment setting results in opposite conclusions.

\section{Ethical Considerations}

Although we showed strong correlations between WEAT and the WinoBias extrinsic bias metrics, researchers and practitioners should not replace the WinoBias extrinsic bias metrics with WEAT.
The purpose of our study is to assess the reliability of intrinsic bias metrics.
Even when a word embedding has the WEAT metric of 0, the WinoBias extrinsic bias metric shows an F1 difference of around 0.1. (the top-left of Figure~\ref{fig:results})
Replacing the WinoBias extrinsic bias metrics with WEAT might make us underestimate the bias.

\section{Limitations}
We could not show why the intrinsic bias metrics correlate with the WinoBias extrinsic bias metrics but not the HSD bias metrics.
Analyzing correlations with more extrinsic bias metrics might be necessary to discover why.

\section*{Acknowledgements}
This paper is written with ``\href{https://www.overleaf.com/latex/templates/association-for-computational-linguistics-acl-conference/jvxskxpnznfj}{the LaTeX template for *ACL conferences}'' by Association for Computational Linguistics, licensed under \href{https://creativecommons.org/licenses/by/4.0/}{CC BY 4.0}.
We thank Associate Professor Hitomi Yanaka for a helpful comment.
We thank anonymous OpenReview reviewers for helpful discussions.

% Bibliography entries for the entire Anthology, followed by custom entries
%\bibliography{anthology,custom}
% Custom bibliography entries only
%\bibliography{custom}

\bibliography{acl_latex}

\appendix

\textbf{Note: The following appendixes contain offensive language.}

\section{Word Sets Extraction} \label{app:word_ext}

We select target and attribute words from the candidates.
The average number of target and attribute words in the corresponding WEAT word sets decides how many target and attribute words are chosen.
When extracting words for gender bias, the corresponding WEAT word sets are WEAT 6, 7, and 8, designed to measure gender bias.
For the same reason, the corresponding WEAT word sets are WEAT 3, 4, and 5 when extracting words for racial bias.

We extracted the following candidate words from the gender HSD dataset used in \citet{goldfarb-tarrant-etal-2021-intrinsic} (Table~\ref{tab:gender_hate_can}).
\begin{table*}
    \centering
    \begin{tabular}{>{\raggedright\arraybackslash}p{2.75cm}>{\raggedright\arraybackslash}p{2.75cm}>{\raggedright\arraybackslash}p{2.75cm}>{\raggedright\arraybackslash}p{2.75cm}}
        \hline
            Male target words & Female target words & Hateful attribute words & Unhateful attribute words\\
        \hline
            he's, boy, “, ”, tomorrow, he, health, guy, past, moment, i'd, bill, him, king, ... &
            women's, babe, lady, almost, movie, beauty, bitches, taking, kendall, jenner, wife, close, HUNDRED POINTS SYMBOL, sweetie, ... &
            flourish, evil, yall, fuckin, slut, females, idiot, niggas, bitches, fuck, bitch, stupid, hate, nigga, ... &
            thanks, win, long, –, liked, top, star, national, health, 7, public, pretty, amazing, 1st, ...\\
        \hline
    \end{tabular}
    \caption{Candidate words extracted from the HSD dataset for gender bias. All capital words represent emojis. We show only the first 14 words here.}
    \label{tab:gender_hate_can}
\end{table*}
We removed ``flourish,'' ``yall,'' and ``females'' from hateful words for not representing hate.
We removed ``slut,'' ``bitches,'' and ``bitch'' for associations with females.
We then chose the first eight words in candidates for unhateful and hateful attribute words.
We chose ``he's,'' ``boy,'' ``he,'' ``him,'' and ``king'' as male target words and ``women's,'' ``lady,'' and ``wife,'' as female target words.\footnote{We did not choose ``guy,'' ``babe,'' and ``bitches'' because their opposite sex counterparts are unclear.}
For each word, we added the word of the opposite gender.
Table~\ref{tab:gender_hate_words} shows the final target and attribute words.
\begin{table*}
    \centering
    \begin{tabular}{>{\raggedright\arraybackslash}p{2.75cm}>{\raggedright\arraybackslash}p{2.75cm}>{\raggedright\arraybackslash}p{2.75cm}>{\raggedright\arraybackslash}p{2.75cm}}
        \hline
            Male target words & Female target words & Hateful attribute words & Unhateful attribute words\\
        \hline
            he's, boy, he, him, king, men's, gentleman, husband &
            she's, girl, she, her, queen, women's, lady, wife &
            evil, fuckin, idiot, niggas, fuck, stupid, hate, nigga &
            thanks, win, long, –, liked, top, star, national\\
        \hline
    \end{tabular}
    \caption{Target and attribute words extracted from the HSD dataset for gender bias} \label{tab:gender_hate_words}
\end{table*}

We extracted the following candidate words from the racial HSD dataset (Table~\ref{tab:race_hate_can}).
\begin{table*}
    \centering
    \begin{tabular}{>{\raggedright\arraybackslash}p{2.75cm}>{\raggedright\arraybackslash}p{2.8cm}>{\raggedright\arraybackslash}p{2.75cm}>{\raggedright\arraybackslash}p{2.8cm}}
        \hline
            White target words & African-American target words & Unhateful attribute words & Hateful attribute words\\
        \hline
            amazing, automatically, anyone, trecru, giveaway, awesome, nice, seeing, power, here's, pick, easter, aren't, america, virgo, series, photos, less, inside, artist, area, THUMBS UP SIGN, weeks, ... &
            goin, ion, bad.can, stans, shorty, females, fa, flourish, hoes, niggas, bruh, nigga, yoongi, bout, hoe, nerves, bitches, itgetsworseshow, ain't, FACE PALM OF FEMALE WITH FITZPATRICK SKIN TYPE 4, yall, FACE PALM OF FEMALE WITH FITZPATRICK SKIN TYPE 5, tf, aint, lil, mad, ugly, protesting, mama, sis, tryna, ... &
            unfollowed, ready, liked, latest, social, –, pretty, excited, light, favorite, far, perfect, easy, public, gemini, following, success, playlist, blue, virgo, ... &
            slut, bad.can, FACE PALM OF FEMALE WITH FITZPATRICK SKIN TYPE 5, asshole, moron, fucks, females, fa, bitches, bitch, niggas, fuckin, nigga, stupid, SPLASHING SWEAT SYMBOL, flourish, goin, idiot, sh, fucking, stressing, fucked, fuck, itgetsworseshow, ugly, idiots, UNAMUSED FACE, FACE PALM OF FEMALE WITH FITZPATRICK SKIN TYPE 4, bullshit, SMILING FACE WITH HORNS, bastard, shorty, ...\\
        \hline
    \end{tabular}
    \caption{Candidate words extracted from the HSD dataset for racial bias}
    \label{tab:race_hate_can}
\end{table*}
To choose white and African-American target words, we ran the mixed-membership demographic-language model developed by \citet{blodgett-etal-2016-demographic}.
It receives a list of word tokens and predicts the proportions of dialects of each race and ethnicity (white, African-American, Hispanic, and Asian).
We took the argmax of these proportions for each candidate word to determine its race or ethnicity.
We chose words in Table~\ref{tab:race_hate_words} as the target and attribute words.
\begin{table*}
    \centering
    \begin{tabular}{>{\raggedright\arraybackslash}p{2.75cm}>{\raggedright\arraybackslash}p{2.75cm}>{\raggedright\arraybackslash}p{2.75cm}>{\raggedright\arraybackslash}p{2.75cm}}
        \hline
            White target words & African-American target words & Unhateful attribute words & Hateful attribute words\\
        \hline
        amazing, automatically, anyone, awesome, nice, seeing, power, here's, easter, aren't, series, photos, less, THUMBS UP SIGN, weeks &
        goin, ion, stans, females, fa, bruh, bout, nerves, ain't, yall, aint, lil, mama, sis, tryna &
        ready, liked, latest, social, –, pretty, excited, light, favorite, far, perfect, easy, public, gemini, following, success, playlist, blue, virgo &
        slut, asshole, moron, fucks, bitches, bitch, fuckin, stupid, idiot, fucking, stressing, fucked, fuck, ugly, idiots, UNAMUSED FACE, bullshit, bastard, shorty\\
        \hline
    \end{tabular}
    \caption{Target and attribute words extracted from the HSD dataset for racial bias} \label{tab:race_hate_words}
\end{table*}

\section{WEAT Words Used in Our Experiment} \label{app:weat}
\subsection{WEAT 3}
The original word sets are taken from \citet{Greenwald_McGhee_Schwartz_1998}.
Some of the African American names do not appear in our processed Twitter corpus.
We removed them.
To make target words equal in size, we removed the same number of names from the European American names randomly.
\begin{itemize}
    \item European American names (target words): Adam Roger Alan Frank Justin Ryan Andrew Jack Greg Paul Peter Amanda Melanie Kristin Nancy Stephanie Ellen Lauren Emily Rachel
    \item African American names (target words): Alonzo Jamel Theo Jerome Leroy Torrance Darnell Lamar Lionel Tyree Deion Lamont Malik Terrence Tyrone Ebony Jasmine Tia Yolanda Yvette
    \item Pleasant attribute words: caress freedom health love peace cheer friend heaven loyal pleasure diamond gentle honest lucky rainbow diploma gift honor miracle sunrise family happy laughter paradise vacation
    \item Unpleasant attribute words: abuse crash filth murder sickness accident death grief poison stink assault disaster hatred pollute tragedy bomb divorce jail poverty ugly cancer evil kill rotten vomit
\end{itemize}

\subsection{WEAT 4}
European and African American names are taken from \citet{Bertrand_Mullainathan_2004}.
As in WEAT 3, we removed some names.
Attribute words are the same as WEAT 3.
\begin{itemize}
    \item European American names (target words): Brendan Greg Brett Jay Allison Anne Carrie Emily Laurie Kristen Meredith Sarah
    \item African American names (target words): Darnell Hakim Jermaine Kareem Jamal Leroy Rasheed Tyrone Aisha Ebony Keisha Kenya
\end{itemize}

\subsection{WEAT 5}
Pleasant and unpleasant attribute words are taken from \citet{Nosek_Banaji_Greenwald_2002a}.
Target words are the same as WEAT 4.
\begin{itemize}
    \item Pleasant attribute words: joy love peace wonderful pleasure friend laughter happy
    \item Unpleasant attribute words: agony terrible horrible nasty evil war awful failure
\end{itemize}

\subsection{WEAT 6}
Career and family words are taken from \citet{Nosek_Banaji_Greenwald_2002a}.
The original WEAT 6 has male and female names as target words.
\citet{goldfarb-tarrant-etal-2021-intrinsic} substituted general male and female words in WEAT 7 for male and female names in WEAT 6 because these names do not appear in WinoBias.
\citet{goldfarb-tarrant-etal-2021-intrinsic} also swapped target and attribute words so that male and female words are attribute words as in WEAT 7 and 8.
The word sets used in our experiment are as follows:
\begin{itemize}
    \item Career target words: executive management professional corporation salary office business career
    \item Family target words: home parents children family cousins marriage wedding relatives
    \item Male attribute words: male man boy brother he him his son
    \item Female attribute words: female woman girl sister she her hers daughter
\end{itemize}

\subsection{WEAT 7}
The word sets are taken from \citet{Nosek_Banaji_Greenwald_2002a}.
The word ``computation'' in the math target words does not appear in our processed Tweeter corpus.
We remove it in the experiment dealing with Twitter data.
To make the target words equal in size, we remove the word ``symphony'' from the arts target words.
(``symphony'' was chosen at random.)
\begin{itemize}
    \item Math target words: math algebra geometry calculus equations computation numbers addition
    \item Arts target words: poetry art dance literature novel symphony drama sculpture
    \item Male attribute words: male man boy brother he him his son
    \item Female attribute words: female woman girl sister she her hers daughter
\end{itemize}

\subsection{WEAT 8}
The word sets are taken from \citet{Nosek_Banaji_Greenwald_2002b}.
\begin{itemize}
    \item Science target words: science technology physics chemistry Einstein NASA experiment astronomy
    \item Arts target words: poetry art Shakespeare dance literature novel symphony drama
    \item Male attribute words: brother father uncle grandfather son he his him
    \item Female attribute words: sister mother aunt grandmother daughter she hers her
\end{itemize}

\section{Artifacts We Used}

Table~\ref{tab:assets1} and \ref{tab:assets2} show artifacts we used, their creators, versions, and licenses or terms.

\begin{table*}
    \centering
    \begin{tabular}{p{4cm}p{4cm}p{2cm}p{4cm}}
        \hline
        Artifacts & Creators & Versions & Licenses or terms\\
        \hline
        Bling Fire & \citet{bling_fire_2021} & 0.1.8 & MIT License\\
        NumPy & \citet{harris2020array} & 1.21.6 & The 3-Clause BSD License\\
        pandas & \citet{the_pandas_development_team_2023_7979740, mckinney-proc-scipy-2010} & 2.0.2 & The 3-Clause BSD License\\
        SciPy & \citet{2020SciPy-NMeth} & 1.10.1 & The 3-Clause BSD License\\
        \href{https://github.com/kentonl/e2e-coref/blob/master/setup_training.sh}{\path{e2e-coref/setup_training.sh}} & \citet{lee-etal-2017-end} & & Apache License 2.0\\
        CPython & \citet{cpython} & 2.7.18, 3.7.15, and 3.8.12 & Python Software Foundation License Version 2\\
        AllenNLP and AllenNLP Models & \citet{gardner2018allennlp} & 2.10.1 & Apache License 2.0\\
        PyTorch & \citet{pytorch} & 2.0.1 & The 3-Clause BSD License\\
        TorchText & \citet{torchtext} & 0.15.2 & The 3-Clause BSD License\\
        \href{https://github.com/nmrksic/attract-repel/blob/master/code/attract-repel.py}{\path{attract-repel.py}} & \citet{mrksic-etal-2017-semantic} & & Apache License 2.0\\
        WinoBias & \citet{zhao-etal-2018-gender} & & MIT License\\
        The abusive tweet dataset & \citet{founta-etal-2022} & 2 & \href{https://zenodo.org/record/3706866}{Restricted access}\\
        \href{https://github.com/seraphinatarrant/embedding_bias}{\path{seraphinatarrant/embedding_bias}} & \citet{goldfarb-tarrant-etal-2021-intrinsic} & & I got a permission from the corresponding author.\\
        The Word Embeddings Fairness Evaluation framework & \citet{wefe2020} & 0.4.1 & MIT License\\
        OntoNotes & \citet{Weischedel2017OntoNotesA} & 5.0 & LDC License\\
        scikit-learn & \citet{scikit-learn} & 0.24.2 & The 3-Clause BSD License\\
        WEAT 3, 4, 5, 6, 7, and 8 & \citet{Caliskan_Bryson_Narayanan_2017} & & I cited papers from which WEAT words came. (Appendix~\ref{app:weat}) \\
        NLTK & \citet{Bird_Loper_Klein_2009} & 3.7 & Apache License 2.0\\
        WikiExtractor & \citet{Wikiextractor2015} & 3.0.6 & GNU Affero General Public License v3.0\\
        Wikipedia & \citet{wikipedia} & latest on 2022-10 & \href{https://www.gnu.org/licenses/fdl-1.3.html}{GFDL} and CC BY-SA 3.0\\
        Gensim & \citet{rehurek_lrec} & 4.2.0 & GNU Lesser General Public License v2.1\\
        Matplotlib & \citet{Hunter:2007} & 3.7.1 & \href{https://github.com/matplotlib/matplotlib/blob/main/LICENSE/LICENSE}{LICENSE}\\
        ArchiveTeam JSON Download of Twitter Stream 2017-04 & \citet{Scott_2017} & &\href{https://archive.org/about/terms.php}{Internet Archive's Terms of Use}\\
        spaCy & \citet{honnibal_spacy_2020} & 3.7.1 & MIT License\\
        The mixed-membership demographic-language model & \citet{blodgett-etal-2016-demographic} & & MIT License\\
        \hline
    \end{tabular}
    \caption{Artifacts we used, their creators, versions, and licenses or terms (Part 1)} \label{tab:assets1}
\end{table*}

\begin{table*}
    \centering
    \begin{tabular}{p{4cm}p{4cm}p{2cm}p{4cm}}
        \hline
        Artifacts & Creators & Versions & Licenses or terms\\
        \hline
        Race and ethnicity data for first, middle, and last names & \citet{rosenman_race_2023} & 9.0 & CC0 1.0\\    
        \hline
    \end{tabular}
    \caption{Artifacts we used, their creators, versions, and licenses or terms (Part 2)} \label{tab:assets2}
\end{table*}

\section{Discussion About the Intended Use of Artifacts We Used}

Internet Archive's Terms of Use, which governs usage of the Twitter data, states, ``Access to the Archive’s Collections ... is granted for scholarship and research purposes only.''
Our use is in line with its intended use.

\section{Documentations of the Artifacts}

\begin{itemize}
    \item Wikipedia
    \begin{itemize}
        \item Language: many languages (We only use the English part.)
        \item Type: encyclopedia
        \item Demographics of the authors: all over the world
    \end{itemize}
    \item OntoNotes 5.0
    \begin{itemize}
        \item Language: English, Mandarin Chinese, Arabic, Chinese (We only use the English part.)
%        \item Data sources: telephone conversations, newswire, newsgroups, broadcast news, broadcast conversation, weblogs, religious texts
        \item Linguistic phenomena: syntactic structure, predicate-argument semantics, word sense disambiguation, coreference, named entity
    \end{itemize}
    \item WinoBias
    \begin{itemize}
        \item Language: English
        \item Linguistic phenomena: coreference
    \end{itemize}
    \item Twitter
    \begin{itemize}
        \item Language: many languages (We only use the English part.)
        \item Type: SNS messages
        \item Demographics of the authors: all over the world
    \end{itemize}
    \item The abusive tweet dataset
    \begin{itemize}
        \item Language: English
        \item Type: SNS messages
        \item Demographics of the authors: all over the world
    \end{itemize}
\end{itemize}

\section{Details of Word Embedding Training} \label{app:word_emb}

Training settings for word embeddings are similar to \citet{goldfarb-tarrant-etal-2021-intrinsic}.
We train Skip-gram word2vec with negative sampling \citep{Mikolov_Sutskever_Chen_Corrado_Dean_2013} and fastText \citep{bojanowski-etal-2017-enriching} word embeddings with 300 dimensions.
The maximum distance between a word and a predicted word is set to 5.
Word embeddings for coreference resolution are trained on Wikipedia, and for HSD, on the Twitter data.
Word embeddings are trained with the NLP library Gensim \citep{rehurek_lrec}.

\subsection{Training on Wikipedia}

\citet{goldfarb-tarrant-etal-2021-intrinsic} trained word embeddings on the latest Wikipedia article dump for coreference resolution.
Since we cannot identify which version they trained on, we train word embeddings on the latest Wikipedia dump.
(The Wikipedia dump was downloaded in October 2022.)

Data preprocessing is almost the same as in \citet{goldfarb-tarrant-etal-2021-intrinsic}.
The downloaded Wikipedia article dump is processed with WikiExtractor \citep{Wikiextractor2015} to generate a single text file.
WikiExtractor outputs documents in the format of
\begin{verbatim}
<doc id="" url="" title="">
...
</doc>
\end{verbatim}
We remove \verb|<doc id="" url="" title="">| and \verb|</doc>| and extract only the contents.
Using \verb|text_to_sentences| from Bling Fire \citep{bling_fire_2021}, we split text into sentences.\footnote{The GitHub repository of \citet{goldfarb-tarrant-etal-2021-intrinsic} mentions that the Wikipedia corpus is preprocessed to have one paragraph per line. We make it have one sentence per line because it is easier to run dataset balancing in this format. The reason we chose \texttt{text\_to\_sentences} is because there is a code using it in \citet{goldfarb-tarrant-etal-2021-intrinsic}'s repository.}
Tokenization is done by \verb|nltk.tokenize.word_tokenize| from the Python package Natural Language Processing Toolkit (NLTK) \citep{Bird_Loper_Klein_2009}.
A word that appears less than or equal to ten times is replaced with a special token \texttt{<UNK>}.\footnote{\citet{goldfarb-tarrant-etal-2021-intrinsic} said they replaced words appearing less than ten times with \texttt{<UNK>}, but the code they provided replaces words appearing less than or equal to ten times.}
Unlike \citet{goldfarb-tarrant-etal-2021-intrinsic}, we lowercase the Wikipedia dump after tokenization to reduce the number of vocabulary.

The final Wikipedia dump is the size of 3121412634 word tokens.

\subsection{Training on Twitter}

For HSD, \citet{goldfarb-tarrant-etal-2021-intrinsic} trained word embeddings on 2019 Twitter data.
However, tweets in the abusive tweet dataset \citet{founta-etal-2022} were collected from 2017-03-30 to 2017-04-09.
We train word embeddings on Twitter data from 2017-04-01 to 2017-04-06 to perform better on HSD.
We train them on the data from 04-01 because the \citet{Internet_Archive}, from which we downloaded Twitter data, offers Twitter data per month.
We use the data until 04-06 to make the data size on par with the data used in \citet{goldfarb-tarrant-etal-2021-intrinsic}.

Data preprocessing is almost the same as in \citet{goldfarb-tarrant-etal-2021-intrinsic}.
They removed retweets from the tweet dataset.
The hashtag symbol (\verb|#|), @-mention, and URL are replaced with special tokens \verb|<HASH>|, \verb|<MENTION>|, and \verb|<URL>|, respectively.
Tweets are tokenized by \texttt{nltk.tokenize.TweetTokenizer} from NLTK \citep{Bird_Loper_Klein_2009}.
Tokens appearing less than ten times are replaced with \verb|<UNK>|.
The difference we make in preprocessing is that we remove tweets withheld in some countries and ones withheld due to DMCA (the Digital Millennium Copyright Act) complaints, as well as retweets.
The final Twitter data is the size of 42429319 word tokens.

\section{The Number of Parameters in the Models Used, the Computational Budget, and Computing Infrastructure Used}

The number of trainable parameters in the coreference resolution system is 1757195.
The number of trainable parameters in the HSD system is 360902.

Here, we only explain computationally heavy operations in our study.

Extracting text data from the Wikipedia article dump takes about 4 hours with 112 CPUs (Intel Xeon Gold 6258R).

Training one word2vec on the Wikipedia article dump takes about 4 hours with 9 CPUs (Intel Xeon Platinum 8360Y).
Training one fastText on the Wikipedia article dump takes about 8 hours with 9 CPUs (Intel Xeon Platinum 8360Y).
We train 90 word embeddings for each embedding algorithm.
In total, training word embeddings on the Wikipedia article dump takes about 1080 hours with 9 CPUs.

Expanding word sets takes around 3 hours with 9 CPUs (Intel Xeon Platinum 8360Y).

Dataset balancing takes around 7.5 hours with 36 CPUs (Intel Xeon Platinum 8360Y).
\textsc{Attract-Repel} takes around 9 hours with 8 GPUs.
Training one coreference resolution model takes about 1.25 hours with 1 GPU.
We train 216 models.

Measuring WEAT takes around 2 hours with 9 CPUs (Intel Xeon Platinum 8360Y).
Evaluating coreference resolution models on WinoBias and the test data of CoNLL-2012 Shared Task \citep{pradhan-etal-2012-conll} takes around 2 hours with 9 CPUs (Intel Xeon Platinum 8360Y).

The GPUs we use are NVIDIA A100.

\section{Basic Statistics of the Datasets for Extrinsic Bias Metrics} \label{app:stats}
Basic statistics of the datasets for extrinsic bias metrics are shown in Table~\ref{tab:data_stat1}, \ref{tab:data_stat2}, and \ref{tab:data_stat3}.

\begin{table}
    \centering
    \begin{tabular}{p{3cm}rrr}
        \hline
        Dataset & \multicolumn{3}{c}{Number of examples}\\
        \hline
        & Train & Val & Test\\
        OntoNotes 5.0 & 75187 & 9603 & 9479\\
        WinoBias Type 1 & & 792 & 792\\
        WinoBias Type 2 & & 792 & 792\\
        The HSD dataset & 76496 & 8500 & 15000\\
        \hline
    \end{tabular}
    \caption{The number of examples in each dataset. An ``example'' here means a tweet for the HSD dataset and a sentence for the others.} \label{tab:data_stat1}
\end{table}

\begin{table}
    \centering
    \begin{tabular}{rrr}
    \hline
    Male-targeting & Female-targeting & Neutral\\
    1837 & 1594 & 11569\\ 
    \hline
    \end{tabular}
    \caption{The number of male-targeting, female-targeting, and neutral tweets in the test set of the HSD dataset.} \label{tab:data_stat2}
\end{table}

\begin{table}
    \centering
    \begin{tabular}{r>{\raggedleft\arraybackslash}p{2.9cm}r}
    \hline
    White-sounding & African-American-sounding & Others\\
    9722 & 1798 & 3480\\
    \hline
    \end{tabular}
    \caption{The number of white-sounding, African-American-sounding, and the other tweets in the test set of the HSD dataset.} \label{tab:data_stat3}
\end{table}

\section{Bias Modification Word Sets} \label{app:bias_modi_wordsets}

\citet{goldfarb-tarrant-etal-2021-intrinsic} used \verb|en_vectors_web_lg| from spaCy \cite{honnibal_spacy_2020} as $v$.
We use \verb|en_core_web_lg| since we cannot download \verb|en_vectors_web_lg|.
\citet{goldfarb-tarrant-etal-2021-intrinsic} used $K = 100$, but we use $K = 10$ to prevent inappropriate words sneaking in.\footnote{For example, we got ``girly'' for male attribute expansion when $K = 100$.}
We remove odd terms from the expanded word sets.
\citet{goldfarb-tarrant-etal-2021-intrinsic} does not mention what constitutes ``odd.''
In our experiment, for white-sounding and African-American-sounding word sets, we use the mixed-membership demographic-language model developed by \citet{blodgett-etal-2016-demographic} to rule out odd terms.
For European-American and African-American names word sets, we use a dataset by \citet{rosenman_race_2023}, containing the first names' race and ethnicity probabilities.
For 136K first names, it provides racial and ethnic probabilities of white, black, Hispanic, Asian, and others.
By taking argmax, we estimate races and ethnicities of expanded European-American and African-American names and remove inappropriate names from the word sets.
We manually decided whether to remove words (or names) not in the demographic-language model (or the name dataset) and words in the other word sets.
We did this removal, considering whether words are unnatural to be in the word set.
(E.g., ``kid'' is unnatural to be in a male word set, and ``purple'' is in a pleasant word set.)
Final expanded word sets are accessible in the supplementary material.

\section{Acknowledgement of AI Use}

We used Code Llama as a coding assistant.
We also used Google Bard to ask questions related to coding.

\section{Measuring the RNSB} \label{app:rnsb}

Our implementation of RNSB is based on an implementation in the Word Embeddings Fairness Evaluation framework (WEFE) \citep{wefe2020}.
Logistic regression of the RNSB is \verb|sklearn.linear_model.LogisticRegression| from scikit-learn \citep{scikit-learn}.
Arguments to \verb|sklearn.linear_model.LogisticRegression| is in Table~\ref{tab:lr_arg}.
\begin{table}
    \centering
    \begin{tabular}{cc}
        \hline
        Argument & Value\\
        \hline
        solver & ``liblinear''\\
        max\_iter & 10000\\
        \hline
    \end{tabular}
    \caption{Arguments to \texttt{sklearn.linear\_model.LogisticRegression}. The other arguments are set to default values.}
    \label{tab:lr_arg}
\end{table}
By default, the RNSB in WEFE holds out 20\% of attribute words as test data.
We disable this behavior so that logistic regression is trained on full attribute words.
We train logistic regression ten times with different random seeds.
The averaged RNSB is used to calculate correlations with extrinsic bias metrics.

\section{Hyperparameter Values of Hate Speech Detector} \label{app:hsd}

We use the same hyperparameter values as in \citet{kim-2014-convolutional}.
However, \citet{kim-2014-convolutional} did not mention the mini-batch sizes of validation and test datasets.
We tested the mini-batch sizes of 1 and 50 (the mini-batch size of a training dataset) and found that an F1 score on the test set is higher with the mini-batch size of 50.
Therefore, we set 50 as the mini-batch sizes.

\section{Preprocessing of the Abusive Tweet Dataset} \label{app:hsd_dataset}

The abusive tweet dataset \citep{founta-etal-2022} labeled each tweet with one of the following four labels: abusive, hateful, spam, and normal.
Referring to the annotation scheme of Shared Task ``SemEval-2019 Task 5: Multilingual Detection of Hate Speech Against Immigrants and Women in Twitter'' \citep{basile-etal-2019-semeval}, we convert the dataset to an HSD dataset by translating abusive and hateful labels to hate speech labels and spam and normal labels to non-hate speech labels.
We select around 10\% of tweets from non-test data as validation data, following \citet{kim-2014-convolutional}.

\citet{goldfarb-tarrant-etal-2021-intrinsic} did not mention the preprocessing of the abusive tweet dataset.
We investigated the preprocessed data they shared and tried to infer how the preprocessing was done.
We preprocess the abusive tweet dataset like we do the Twitter data for word embedding training, although we do not filter out tweets.
Words not in the word embedding vocabulary are replaced with \verb|<UNK>|, except ones that appear more than or equal to ten times in the training section of the abusive tweet dataset.

\section{Modifications to Packages}

\subsection{WEFE}

\begin{itemize}
    \item Fix comments and error messages
    \item Handle multiple token expression
    \item When measuring the RNSB, the original WEFE raises an exception if the number of positive training examples is 1. Probably, this is a mistake for ``if the number of positive training examples is less than 1.'' We fixed it.
    \item Fix a docstring
    \item Fix WEAT words
    \item Implement an extension of the RNSB
\end{itemize}

\subsection{AllenNLP and AllenNLP Models}

When measuring the WinoBias score, we use modified versions of AllenNLP and AllenNLP Models.
We relaxed the specifications in the requirement files so that the packages are installed successfully.
We modified AllenNLP Models to handle WinoBias.

\subsection{The Mixed-Membership Demographic-Language Model}

We adapted the mixed-membership demographic-language model by \citet{blodgett-etal-2016-demographic} to Python 3.

\subsection{WikiExtractor}

When we ran WikiExtractor, an exception was raised.
We searched GitHub issues for the same problem and applied a fix found there.

\section{The Combinations of Parameters of \textsc{Attract-Repel}} \label{app:ar_paras}

\textsc{Attract-Repel} \citep{mrksic-etal-2017-semantic} has following parameters: $\delta_\text{sim}$, $\delta_\text{ant}$ (``the similarity and antonymy margins''), $\lambda$ (regularization coefficient), $k_1^\prime$, and $k_2^\prime$ (mini-batch sizes for synonym and antonym pairs).
\citet{mrksic-etal-2017-semantic} conducted a parameter search in the space of $\delta_\text{sim}, \delta_\text{ant} \in \{0, 0.1, 0.2, ..., 1\}$, $\lambda \in \{10^{-3}, ..., 10^{-10}\}$, and $k_1^\prime, k_2^\prime \in \{10, 25, 50, 100, 200\}$.
They reported that $\delta_\text{sim} = 0.6$, $\delta_\text{ant} = 0$, $\lambda = 10^{-9}$, and $k_1^\prime = k_2^\prime = 10, 25, 50$ achived the best result.

When \citet{goldfarb-tarrant-etal-2021-intrinsic} modified word embeddings, they used the best parameter setting with $k_1^\prime = k_2^\prime = 50$.

Taking those into account, we chose the following combinations of the parameters: $\delta_\text{sim} \in \{0, 1\}$, $\delta_\text{ant} \in \{0, 1\}$, $\lambda \in \{10^{-1}, 5 \times 10^{-2}, 10^{-2}\}$, and $k_1^\prime, k_2^\prime = 50$.\footnote{We found that $\lambda = 10^{-3}$ did not yield much debiasing (and over-biasing) effect compared to $\lambda = 10^{-2}$. Therefore, we only consider $\lambda \ge 10^{-2}$.}

\section{Dataset Anonymization / Offensive Content Removal}

When preprocessing datasets, we mostly follow \citet{goldfarb-tarrant-etal-2021-intrinsic}.
To make our experiment close to theirs, we do not do additional dataset anonymization or offensive content removal.

\subsection{Twitter Data}

An @-mention is replaced with a special token \verb|<MENTION>|.

\subsection{The Abusive Tweet Dataset}

The abusive tweet dataset from \citet{founta-etal-2022} contains offensive content.
Since we train hate speech detectors, we do not filter out offensive content.

An @-mention is replaced with a special token \verb|<MENTION>|.

\end{document}